\documentclass[pmlr,twocolumn]{jmlr}
   
\usepackage{longtable}

\usepackage{booktabs}
\usepackage[load-configurations=version-1]{siunitx} 

\theorembodyfont{\upshape}
\theoremheaderfont{\scshape}
\theorempostheader{:}
\theoremsep{\newline}

\jmlrvolume{193}
\firstpageno{1}

\jmlryear{2022}
\jmlrworkshop{Machine Learning for Health (ML4H) 2022}

\usepackage{multirow}
\usepackage{float}
\usepackage{adjustbox}
\usepackage{listings}

\newcommand{\seen} {{\it seen}}
\newcommand{\unseen}{{\it unseen}}

\newcommand{\eg}{\emph{e.g.}}

\DeclareMathOperator*{\argmax}{arg\,max}

\title[OSLAT]{OSLAT: Open Set Label Attention Transformer for Medical Entity Retrieval and Span Extraction}

\author{}
  \author{\Name{Raymond Li}\thanks{work done as a research intern at Curai} \Email{raymondl@cs.ubc.ca}\\
  \addr University of British Columbia
  \AND
  \Name{Ilya Valmianski} \Email{ilya@curai.com}\\
  \addr Curai
  \AND
  \Name{Li Deng}\thanks{work done while at Curai} \Email{l.deng@ieee.org}\\
  \addr Vatic Investments
  \AND
  \Name{Xavier Amatriain} \Email{xavier@curai.com}\\
  \addr Curai
  \AND
  \Name{Anitha Kannan} \Email{anitha@curai.com}\\
  \addr Curai
 }

\begin{document}

\maketitle

\begin{abstract}
\label{sec:abstract}

Medical entity span extraction and linking are critical steps for many healthcare NLP tasks. Most existing entity extraction methods either have a fixed vocabulary of medical entities or require span annotations. In this paper, we propose a method for linking an open set of entities that does not require any span annotations. Our method, \textbf{Open Set Label Attention Transformer (OSLAT)}, uses the label-attention mechanism to learn candidate-entity contextualized text representations. We find that OSLAT can not only link entities but is also able to implicitly learn spans associated with entities. We evaluate OSLAT on two tasks: (1) span extraction trained without explicit span annotations, and (2) entity linking trained without span-level annotation. We test the generalizability of our method by training two separate models on two datasets with low entity overlap and comparing cross-dataset performance.
\end{abstract}

\section{Introduction}
\label{sec:intro}

Many natural language processing (NLP) tasks in the healthcare domain such as information retrieval (IR) \citep{10.1145/3462476}, diagnosis coding \citep{early-icd10-classification}, and conversational agents \citep{MEDCOD, pmlr-v158-valmianski21a} greatly benefit from correctly identifying medical entities such as disorders and findings in the text. This has led to a wealth of literature centered on entity recognition in the past decades \citep{Fries2020, pmid:8563299, pmid:12123149, metamap, savova2010mayo} and many competitions/ tasks in both NLP and IR communities \citep{hNLP4, hNLP1, hNLP3, hNLP2}.

\begin{table*}
\centering

\begin{tabular}{ll}
\toprule
\textbf{Entity}   & \textbf{Text containing the entity}\\ 
\midrule
\vspace{5pt}
\begin{tabular}[c]{@{}l@{}}
knee swelling
\end{tabular}  &
\begin{tabular}[c]{@{}l@{}}
pain and 
\colorbox{green!30.98630142211914}{swelling} \colorbox{green!28.580547332763672}{in} \colorbox{green!30.98630142211914}{knee}
\end{tabular}  
\\ 
\vspace{5pt}
\begin{tabular}[c]{@{}l@{}}
knee pain
\end{tabular}  &
\begin{tabular}[c]{@{}l@{}}

\colorbox{green!30.98630142211914}{pain}{and swelling}  \colorbox{green!28.580547332763672}{in}\colorbox{green!30.98630142211914} {knee}

\end{tabular}  
\\ 
\vspace{5pt}
\begin{tabular}[c]{@{}l@{}}
cervical \\ lymphadenopathy
\end{tabular}  &
\begin{tabular}[c]{@{}l@{}}

\colorbox{green!28.580547332763672}{swollen} \colorbox{green!30.98630142211914}{lymph} \colorbox{green!7.59257698059082}{node} \colorbox{green!0.0}{on} \colorbox{green!0.0}{right} \colorbox{green!0.0}{side} \colorbox{green!0.0}{of} \colorbox{green!38.1240234375}{neck}

\end{tabular}  
\\ 
\begin{tabular}[c]{@{}l@{}}
dyspnea
\end{tabular}  &
\begin{tabular}[c]{@{}l@{}}

\colorbox{green!0.0}{head pressure and anxiety for the past couple weeks also,} \\
\colorbox{green!0.0}{having}
\colorbox{green!1.2714784145355225}{to} \colorbox{green!11.428217887878418}{take} \colorbox{green!9.253361701965332}{really} \colorbox{green!14.602005004882812}{deep} \colorbox{green!10.144306182861328}{breaths} \colorbox{green!0.0}{to} \colorbox{green!24.001874923706055}{catch} \colorbox{green!0.0}{my} \colorbox{green!15.694615364074707}{breath} \colorbox{green!0.0}{.}
\end{tabular} 
\\
\bottomrule
\end{tabular}

\caption{We can see that the entity can present as a contiguous-span of text (row 1), disjoint-spans (rows 2-3) or overlapping-spans (row 1-2). For each (text, entity), the color saturation highlights the prediction confidence from OSLAT.}
\label{table:example-annot}
\vspace{-1.5em}
\end{table*}

However, the problem of entity recognition continues to be largely unsolved, with two main challenges being: (1) lack of sufficient amounts of labeled and diverse data and (2) the ability to handle previously unseen entities (open-set recognition). 

Existing methods require significant amounts of labeled data \citep{PMID:30617335} and often include two parts: (1) entity span annotations, and (2) span-entity linking annotations. For formal clinical texts, such as medical literature and physician notes, weak-labeling approaches (\eg~lookup-based) help reduce the need for span annotations \citep{Fries2020}. However, these approaches struggle with patient-derived text due to insufficient vocabulary coverage and the propensity of patient text to have disjoint spans for entities (see \autoref{table:example-annot}). 

The in-the-wild open-set recognition challenge appears when the models are exposed to text containing entities not seen during training \citep{DBLP:journals/corr/abs-1910-02830, Mottaghi20}. Even when using UMLS \citep{UMLS} as the basis for medical vocabulary, it is difficult to collect enough data to cover \emph{all} entities. Furthermore, real-life text often contains medically relevant compositional entities (\eg~``severe sudden abdominal pain") and colloquial language not in UMLS.


In this paper, we tackle both challenges through \textbf{Open Set Label Attention Transformer (OSLAT)}. OSLAT is similar, and similarly computationally efficient, to a bi-encoder information retrieval architecture. However, unlike bi-encoders, it uses the label-attention mechanism to create candidate-entity contextualized document representations, which can then be used to classify the presence of the candidate entity. Like bi-encoders, OSLAT can link open set entities and does not require span annotations. However, the label attention mechanism allows it to implicitly learn to infer entity span masks, even for disjoint spans. 


We summarize our work by outlining the two technical contributions:
\begin{enumerate}
    \item We use a transformer-based encoder to encode not only the input text but also the candidate labels in a label-attention architecture. This allows us to operate on an open set of labels.
    \item  To train our model, we introduce Label Synonym Supervised Normalized Temperature-Scaled Cross-Entropy (LSS-NT-Xent) loss, an extension of NT-Xent \citep{chen2020simple}
\end{enumerate}
We test the generalizability of our approach by performing extensive experiments on two datasets. 
Including evaluating on unseen entities, and applying a model trained on one dataset to a test set in the other dataset.\footnote{Our implementation publicly is available. at:\\\href{https://github.com/curai/curai-research/tree/main/OSLAT}{https://github.com/curai/curai-research/tree/main/OSLAT}}

\section{Definition and Tasks Formulations}
\label{sec:definition}
We begin by defining a universe of all entities, denoted by $\mathcal{E}$. Note, we do not need to explicitly define $\mathcal{E}$. During training, we will observe a subset of these entities $\mathcal{E}_{\seen}$  and the remaining unobserved (open-set) is $\mathcal{E}_{\unseen} =\mathcal{E} \setminus \mathcal{E}_{\seen}$. We then assume access to a dataset $\mathcal{D}_{train} = \{({\bf x}_t,e_t)\}_{t=1}^{T}$, where ${\bf x}_t$ is the $t^{th}$ target text and $e_t$ is an entity present in it, with $\mathcal{E}_{\seen} = \cup_{t=1}^{T}{e_t}$. Note that  entity mentions/spans are not available during training. For each entity, ${e_t} \in \mathcal{E}_{\seen}$, we also assume access to its synonyms, obtained from an external source such as UMLS (which we use in this paper). 
\paragraph{Task 1: Entity Span Extraction.}
For entity span extraction, we are provided with input text-entity pair $({\bf x}, e)$ s.t. $e \in \mathcal{E}$, this reflects the application of the model in the wild. The goal of this is to identify the spans of text in ${\bf x}$ that describe $e$. 
\paragraph{Task 2: Entity Linking.}
For entity linking, we are provided with the input text and a finite universe of entities $\mathcal{E}_{test} \subseteq \mathcal{E}$. The goal of this task is to predict whether entity $e$ is mentioned in text ${\bf x}$, denoted as $s(x, e)$ s.t. $e \in \mathcal{E}_{test}$, which we refer to as the retrieval score.


\section{Approach}
\label{sec:approach}
\begin{figure*}[ht!]
\centering
\includegraphics[width=\textwidth]{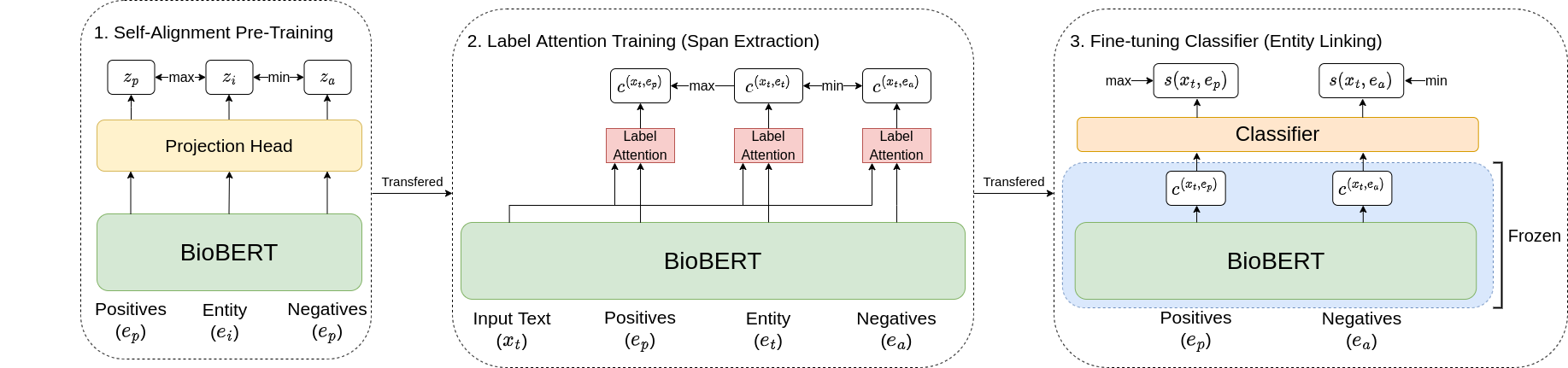}
\caption{Overview of our proposed two-stage training approach. We first perform self-alignment pretraining on medical entities (\S~\ref{sec:approach_pretrain}), before aligning the label-text joint representations obtained through label attention (\S~\ref{sec:lbltransformer}). Finally, we train a entity linking classifier on top of the frozen representations (\S~\ref{sec:classifier-entity-linking}).
}
\label{fig:approach}
\vspace{-1em}
\end{figure*}

Our method, \textbf{Open Set Label Attention Transformer (OSLAT)}, consists of separately encoding the entity and  the text using a single encoder with the representations combined using label attention. In particular, the mean pooled representation of the entity is used to construct a query vector while the token embeddings of the text are used to construct the key-value vector pairs. This query, keys, and values are then used in a softmax attention to compute a single vector (OSLAT representation). The spans are inferred by looking at the label attention scores over the text tokens. This architecture is trained in two stages (see \autoref{fig:approach} for an overview). In the first stage, we perform self-alignment training of the encoder (\S~\ref{sec:approach_pretrain}). In the second stage, we use our extension of NT-Xent, which we call LSS-NT-Xent, to contrastively train the OSLAT representations (\S~\ref{sec:lbltransformer}). Finally, we perform entity linking by training a binary classifier on top of the OSLAT representations (\S~\ref{sec:classifier-entity-linking}).


\vspace{-0.5em}

\subsection{Self-Alignment Pretraining on Medical Entities}
\label{sec:approach_pretrain}

We  use BioBERT \citep{lee2020biobert} as the backbone encoder. We perform self-alignment pretraining to decrease representational anisotropy of entity embeddings \citep{li-etal-2020-sentence,Carlsson2021, gao-etal-2021-simcse, liu-etal-2021-self, liu-etal-2021-mirror-bert} (for the change in anisotropy see \autoref{fig:entity_density} in Appendix \ref{sec:entity-density}). In particular, for medical entity $e_i \in \mathcal{E}_{\seen}$, we obtain its representation $h^{(e_i)}$ by taking the $\textsf{[CLS]}$ token embedding of the last hidden layer of BioBERT. To apply the contrastive loss function, we follow the model architecture described in SimCLR \citep{chen2020simple}, where a two-level feed-forward projection head maps the representation $h^{(e_i)}$ from BioBERT into a low-dimension space, before a supervised contrastive loss, NT-Xent, is applied to the normalized projection output $z_{i}$ \citep{chen2020simple, khosla2020supervised, gao-etal-2021-simcse}:
\begin{align}
\begin{split}
\mathcal{L}_{\textrm{pre}} = \sum_{i \in B} &\frac{-1}{|P(i)|} \\
\Bigg( \sum_{p \in P(i)} &\log\frac{\exp(z_i \cdot z_p / \tau)}{\sum_{a \sim \mathcal{E}_{\seen}}\exp(z_i \cdot z_a / \tau)} \Bigg)
\end{split}
\end{align}

For each entity in batch $B$, the positives $z_p$ are projected representations from the synonym set $P(i)$ of entity $e_i$, with $|P(i)|$ as its cardinality, while the negatives $z_a$ are projected representations from sampled entities from $\mathcal{E}_{\seen}$. Finally, hyperparameter $\tau$ denotes the scalar temperature. As the entities are organized into disjoint synonym sets, we apply a stratified sampling strategy for sampling negatives, where we first sample a synonym set and then sample an entity from that set. This ensures that entities with a smaller synonym set do not get under-represented during training.
After the self-alignment pretraining, we discard the projection head keeping the fine-tuned encoder. Details on our training procedure can be found in \S~\ref{sec:setup}.

\vspace{-.5em}
\subsection{Label Attention Training}
\label{sec:lbltransformer}
OSLAT supports an open set of labels by jointly encoding labels and target texts into the same subspace. To obtain the representation of the entity spans within the target text, we first encode label $e_t$ and target text ${\bf x}_t$ with our self-alignment pretrained BioBERT (see \S~\ref{sec:approach_pretrain}). Specifically, for $({\bf x}_t,e_t) \in \mathcal{D}$, the label representation $h^{(e_t)} \in \mathbb{R}^{1 \times d}$ and target text representation $h^{(x_t)} \in \mathbb{R}^{n \times d}$ from the last hidden layer of BioBERT (with hidden size $d$) are used to compute the label-attention score using a variant of the dot-product attention:

\vspace{-1em}
\begin{equation}
    \alpha^{({\bf x}_t,e_t)} = \textrm{Softmax}\big(h^{(e_t)} (h^{(x_t)})^T\big)
\end{equation}

where the attention score $\alpha^{({\bf x}_t,e_t)}_k$ can be interpreted as the token-wise semantic similarity between the label $e_t$ and the $k$th token of target text ${\bf x}_t$. Since the $\textsf{[CLS]}$ token for the target text can contain aggregate semantic information about the entire input, we found that the model often resorted to attending solely to the $\textsf{[CLS]}$ token. To mitigate this issue, we remove the $\textsf{[CLS]}$ token from $h^{(x_t)}$ to encourage the model to attend to other portions of the target text. Finally, we compute the entity span representation as a weighted sum of the target text $h^{(x_t)}$ by the attention scores:
\begin{equation}
    c^{({\bf x}_t,e_t)} = \sum_{k=1}^n \alpha^{({\bf x}_t,e_t)}_k h^{(x_t)}_k
\end{equation}
To train our model, we use a  variant of NT-Xent which we call Label Synonym Supervised Normalized Temperature-Scaled Cross Entropy (LSS-NT-Xent):

\begin{align}
\begin{split}
    &\mathcal{L}_{\mathrm{LSS}}(I) = \sum_{t \in B}\frac{-1}{|P(t)|} \\
    \Bigg( &\sum_{p \in P(t)}\log\frac{\exp(c^{({\bf x}_t,e_t)} \cdot c^{({\bf x}_t,e_p)} / \tau)}{\sum_{a \sim \mathcal{E}_{\seen}} \exp(c^{({\bf x}_t,e_t)} \cdot c^{({\bf x}_t,e_a)} / \tau)} \Bigg)
\end{split}
\end{align}
Similar to the self-alignment pre-training described in \S~\ref{sec:approach_pretrain}, we use $e_t$'s synonym set $P(t)$ as positives and randomly sample negatives from the $\mathcal{E}_{\seen}$ and their synonyms. 
At inference time, we use the attention scores $\alpha^{({\bf x}_t,e_t)}$ to predict whether each token of $x_t$ lies in the span of entity $e_t$.

\subsection{Entity Linking}
\label{sec:classifier-entity-linking}
For the task of entity linking, we apply a binary classifier on top of the entity span representation $c^{(x_t, e)}$ to predict the probability of entity $e$ being mentioned in text $x_t$. During training, we optimize the classifier with Focal Loss \citep{lin2017focal}, where the parameters of the OSLAT encoder are kept frozen to ensure that its attention weights can still be used for span extractions. For each example, the positives are synonyms of mentioned entities, while negatives are sampled from the universe of all entities. In practice, training time can be significantly reduced by caching the entity span representations of the training set in the first epoch.

At inference time, we use the classifier output as the unnormalized retrieval score:
\begin{equation}
    s(x_t, e) = \textrm{Classifier}(c^{(x_t, e)})
\end{equation}

\section{Related Work}
\label{sec:related_work}

\paragraph*{Entity mention/Span detection.} Unlike OSLAT, most approaches ({\it c.f.} \citet{10.1007/978-981-16-2597-8_13} for a comprehensive survey) for this task require access to a manually-labeled span-level dataset during training. Notable exceptions include \citet{Fries2020} that uses weak supervision to construct a labeled training set for the task or \citet{Fu2021} that learns to reconcile outputs from multiple span detection approaches. However, models used to generate inputs to \citet{Fu2021} require manual annotations.

\paragraph*{Entity linking.} Most methods ({\it c.f.} \citet{DBLP:journals/corr/abs-2006-00575} for a survey) require datasets with explicit spans during training, and often operate within a closed set of entities. While \citet{Wu2019, Mottaghi20} operates with an open set, \citet{Wu2019} requires labeled entity spans while \citet{Mottaghi20} sidesteps entity mention problem and poses it as a classification task within active learning paradigm. In contrast, OSLAT takes as input an entity of interest, and then simultaneously detects a span of text and whether the entity of interest is in that span. 


\paragraph*{Label attention in healthcare.}

Label attention for classification over a fixed label set is studied in various healthcare applications \citep{mullenbach-etal-2018-explainable, Vu2020, liu-etal-2021-effective, squeeze-and-excite, LATA, LAME}. 

Most previous approaches used it within a convolutional \citep{mullenbach-etal-2018-explainable, liu-etal-2021-effective, squeeze-and-excite}  or BiLSTM \citet{Vu2020} architecture for the task of classifying clinical notes into International Classification of Diseases (ICD) codes. Meanwhile, some recent works have also used transformer architecture \citep{LATA, LAME} for this task. In particular, \citet{LAME} extends \citet{LATA}  by incorporating the label attention module into the encoder fine-tuning process, leading to improvements in the encoder representations of biomedical text. 


However, these previous approaches focus on the problem of multi-label classification over a fixed label set. In contrast, OSLAT can infer disjoint spans in the input text that map to an entity within an open set. To the best of our knowledge, this is the first work to address this problem in an open-set context.


%

\section{Datasets}
\label{sec:dataset}

We are interested in investigating the following empirical questions: 
\begin{itemize}
    \item \textbf{Open set entity detection}: \\Is our approach robust to entities that are unseen during training?
    \item \textbf{Cross-domain transfer}: \\Is our approach robust when applied to data from a different domain than that of training
    (\eg~train on patient written and apply to provider/expert-written text)?  
    \item \textbf{Handling disjoint-spans}: \\Is our approach robust in identifying entity spans that are disjoint? 
\end{itemize}

In order to answer these questions, we build two complementary datasets. The first dataset (\S~\ref{sec:rfe-dataset}) is comprised of texts in which patients describe their health issues (RFE dataset). The second dataset (\S~\ref{sec:hnlp-dataset}) is comprised of discharge summary notes written by physicians (hNLP dataset). The train-test split procedure (\S~\ref{sec:dataset_construct}) of these datasets is itself non-trivial as we need to split both target texts and medical entities such that the test set contains both  \seen~and \unseen~ entities. Lastly, it's worth mentioning that there is a significant difference between the entity sets in both datasets (85\%  from hNLP to RFE and 69\% from RFE to hNLP), this ensures that we do not provide undue advantage to the model during cross-domain evaluations. A detailed comparison between the two datasets is available in Appendix \ref{sec:dataset_comparison}.

\begin{table*}[ht!]
\centering
\begin{tabular}{@{}cccccccc@{}}
\toprule \vspace{4pt}
 &  & \multicolumn{2}{c}{\textbf{Seen}} & \multicolumn{2}{c}{\textbf{Unseen}} & \multicolumn{1}{c}{\textbf{Disjoint-Spans} } \\
 &  & $\|\mathcal{E}_{\textnormal{seen}}\|$ & \# Examples &  $\|\mathcal{E}_{\textnormal{unseen}}\|$ & \# Examples & Fraction of examples \\\midrule
\multirow{2}{*}{\textbf{RFE}} & Train & 450 & 6430 & n/a & n/a & unk \\
 & Test & 73 & 266 &  66 &  863 & 13\%\\
\multirow{2}{*}{\textbf{hNLP}} & Train & 1054 & 4377 & n/a & n/a  & 5\%\\
 & Test & 61 & 185 & 143 & 1018 & 7\% \\
 \bottomrule
\end{tabular}
\caption{Statistics of the datasets used in our experiments. 
Note that we do not need access to spans during training and therefore did not obtain span-level annotations for the RFE training set.}
\label{tab:data_stats}
\end{table*}

\subsection{Train/Test dataset construction}
\label{sec:dataset_construct}

We start with an intermediate dataset of the form (${\bf x}_k, E_{k}$) where ${\bf x}_k$ is the $k^{th}$ input text that has a set $E_k$ of entities to reflect that multiple entities can be in the same input text. Then, $\mathcal{E} = \cup_{k}  E_k$ is the universe of entities, and $p(e)$ is the marginal probability of entity $e$ in the dataset. 
\paragraph*{Constructing $\mathcal{E}_{\seen}$, $\mathcal{E}_{\unseen}$:} For our experiments,  we choose 10\% of the entities as \unseen. We choose these entities randomly from 20\%, 40\%, and 40\%  from high, medium, and low marginal probability bins of $p(e)$ so that we capture entities across the spectrum of frequency distribution.

\paragraph*{Train-Test split:} We split the dataset into disjoint sets for training and testing from the perspective of the entity. For each entity $ e \in \mathcal{E}_{unseen}$, we associate all pairs ($({\bf x}_k,e)_{k: {e \in E_k}} $) to the test set.  For each entity $ e \in \mathcal{E}_{seen}$, we  randomly sample, without replacement, 10\% of  {${\bf x}_k,e)_{k: {e \in E_k}} $} pairs for the test set and remaining 90\% to training set. We ensure that all entities in $\mathcal{E}_{seen}$ have at least five examples in the training set. If not, we first prioritize adding to the training set.

\paragraph*{Span level labels for test set:} We also augment the test set with the spans that correspond to the concept. In particular, an example in the test set is of the form (${\bf x},e, { \{{\bf s}_{i,e}\}} $)  where ${ \{{\bf s}_{i,e}\}}$ is the set of spans that collectively identify the entity $e$ in the text ${\bf x}$. In particular, each element in  ${ \{{\bf s}_{i,e}\}}$  encodes the character level beginning and end of the phrase in ${\bf x}$ that is constituent of $e$.

Thus, $\mathcal{D}_{train} = \{({\bf x}_t,e)\}_{t=1}^{T}$ where $e \in   \mathcal{E}_{\seen}$ and  $\mathcal{D}_{test} = \{({\bf x}_k,e,{ \{{\bf s}_{i,e}\})} \}_{k=1}^{K}$, where $e \in  \mathcal{E}$.

\subsection{Dataset 1: Reason for Encounter}
\label{sec:rfe-dataset}
The Reason for Encounter (RFE) dataset is gathered from a telemedicine practice. Patients starting a visit describe their reason for seeking an encounter. This dataset contains a labeled subset of 4909 encounters with 4080 patients. The language used in RFE is more colloquial and less standardized, featuring many disjoint spans for medical entities. Each RFE is labeled by medical experts with corresponding medical findings using UMLS ontology. The RFE examples have an average length of 26 words. 

We constructed the train-test dataset as outlined in \S~\ref{sec:dataset_construct}. In particular,   $|\mathcal{E}_{\seen}| = 450$  and  $|\mathcal{E}_\unseen| = 73$. This results in roughly 90\% of the RFE dataset having at least one entity that is \seen. Further, 24\% of the examples in the RFE dataset have at least one entity in $\mathcal{E}_\unseen$, and 10\% of examples in the RFE dataset have all their entities in $\mathcal{E}_\unseen$. We also provide the demographic breakdown of this dataset in Appendix \ref{sec:rfe-details}.



\vspace{-.5em}
\subsection{Dataset 2: hNLP dataset}
\label{sec:hnlp-dataset}
\vspace{-0.5em}
Our second dataset is derived from the training data in the SemEval-2015 Task 14 \citep{hNLP3}. In particular, we start with the provided 136 discharge notes and their corresponding medical concepts along with their location spans. We split each discharge note into smaller text chunks using the newline delimiter. We removed chunks that do not have any entities associated with them. This leads to 5508 text chunks with an average length of 69.08 words. We built an initial dataset with text chunks, their entities, and their spans. These entities are UMLS Concept Unique Identifiers (CUIs).

We then constructed the train-test dataset as outlined in \S~\ref{sec:dataset_construct}.   $|\mathcal{E}_{\seen}| = 1054$  and  $|\mathcal{E}_{\unseen}| = 143$. This results in roughly 90\% of the examples having at least one entity that is \seen. For more detailed statistics on the dataset, see \autoref{tab:data_stats}. For all examples in the test set, we attach the corresponding spans provided in the original dataset. We do not use these spans during training.

\vspace{-1em}
\section{Task 1: Span Extraction}
\label{sec:span-extract}

In this section, we describe the experiments for the task of entity span extraction using the OSLAT model described in \S\ref{sec:lbltransformer}.

\begin{table*}[ht!]
\begin{tabular}{@{}lcccc@{}}
\toprule
\multicolumn{1}{c}{\multirow{3}{*}{Dataset}} & \multicolumn{2}{c}{\textbf{hNLP}} & \multicolumn{2}{c}{\textbf{RFE}} \\
\multicolumn{1}{c}{} & Contiguous-Span & Disjoint-Span & Contiguous-Span & Disjoint-Span \\
\multicolumn{1}{c}{} & s/u/all & s/u/all & s/u/all & s/u/all \\ \midrule
\textbf{Rule-Based} & - / - / .74 & - / - / .41 & - / - / .55 & - / - / .23 \\
\textbf{Fuzzy-Match} & - / - / \textbf{.79} & - / - / .30 & - / - / .35 & - / - / .20 \\
\textbf{OSLAT} & .77/\textbf{.73}/.74 & - /\textbf{.47}/.47 & \textbf{.67}/\textbf{.59}/\textbf{.66} & \textbf{.56}/\textbf{.60}/\textbf{.57} \\
\begin{tabular}[c]{@{}l@{}}\textbf{OSLAT (CD)}\end{tabular} & \textbf{.80}/.65/.69 & \textbf{.56}/.43/\textbf{.53} & .63/.52/.57 & .52/.41/.45 \\
\begin{tabular}[c]{@{}l@{}}\textbf{OSLAT (NP)}\end{tabular} & .02/.02/.02 & .00/.00/.00 & .12/.11/.12 & .05/.03/.05 \\ \bottomrule
\end{tabular}
\caption{Micro-F1 scores for entity span extraction on both datasets, broken down by spans as well as examples with \seen~(s) and \unseen~(u) entities. We do not report separate seen and unseen values for the baselines since they are provided ground truth entities during inference. All reported results are averaged across 3 seeds, with $\sigma \leq 0.01$.}
\label{tab:results-main}
\vspace{-2em}

\end{table*}

\subsection{Set-up}
\label{sec:setup}
For entity span extraction, we compute the entity-attention scores for the ground-truth entities present in each input text. For experiments on both datasets, 
we compute the average entity-attention scores across all synonym terms associated with each ground-truth entity (identified by a UMLS CUI) as the exact matching synonym is not provided in the annotation. 
Since the attention scores are normalized through the softmax operation (sum up to 1), we manually set the threshold to be $0.05$ during inference. Lastly, we also remove stop-words and punctuation marks from the predictions. 

\subsection{Metrics} 
We use the per-token micro-F1 score as the primary metric for evaluating our models for entity span extraction. This is done by computing the per-token precision and recall based on the token overlaps between the predicted and ground-truth spans before averaging across all examples. 

\subsection{Baselines}  
\label{sec:span-baselines}

We compare \textbf{OSLAT} with the following baselines:
\begin{enumerate}
    \item \textbf{Rule-based.} This is an in-house developed lookup-based approach that uses a sliding window strategy to find maximal matches of text corresponding to the entities and their synonyms. It ignores stop words while doing the match. 
    \item \textbf{Fuzzy-Match.} We adopt the fuzzy-string matching from the implementation by RapidFuzz \citep{max_bachmann_2021_5584996}, where spans with normalized Levenshtein Distance \citep{Levenshtein1966} greater than a threshold are extracted for each entity. 
    \item \textbf{OSLAT (NP).} Ablation of OSLAT without self-alignment pretraining. 
    \item \textbf{OSLAT (CD).} Cross-dataset evaluation (model trained on RFE while evaluated on hNLP and vice versa). 
\end{enumerate}

Rule-based and fuzzy-match baselines are particularly strong because they are provided with the target entity and only need to string match one of the known entity synonyms to the target text. In particular, we find that these two baselines have very high precision, since the matched synonym is almost always the correct span.

\subsection{Results}
\label{sec:crossdomain-results}

\autoref{tab:results-main} shows the micro-F1 scores. We find that on the more challenging RFE dataset, OSLAT achieves the best performance. Even on the hNLP dataset, the RFE-trained OSLAT (OSLAT CD for hNLP) performs best on disjoint-spans. We believe that this is because the higher number of disjoint and otherwise complicated spans (where the entity synonyms do not directly match the span text) in the RFE dataset force the model to learn more abstract label-attention representations. Note that this also demonstrates the generalizability of OSLAT, as there is a low entity overlap between the two datasets (Appendix \ref{sec:dataset_comparison}).

For both datasets, we also find that our model performs well for entities unseen during training. 
Since the synonym set often contains paraphrases of the same entity (\eg~stuffy nose, clogged nose), we hypothesize that our model learns to interpolate within the entity representation space and generalize to paraphrases for unseen entities. 



\begin{table*}[ht!]
\small
\begin{tabular}{@{}lcccc@{}}
\toprule
\multicolumn{1}{c}{\multirow{2}{*}{Dataset}} & \multicolumn{2}{c}{\textbf{hNLP}} & \multicolumn{2}{c}{\textbf{RFE}} \\
\multicolumn{1}{c}{} & Contiguous-Span & Disjoint-Span & Contiguous-Span & Disjoint-Span \\
Accuracy & @1/@5/@10 & @1/@5/@10 & @1/@5/@10 & @1/@5/@10 \\ \midrule
\textbf{Fuzzy-Match} & \textbf{.289/.731/.840} & .018/.140/.228 & .482/.740/.784 & .123/.477/.554 \\
\midrule
\textbf{BioBERT (Unsup)} & .080/.134/.168 & .070/.105/.211 & .186/.031/.352 & .077/.154/.185 \\
\textbf{BioBERT (MS)} & .173/.280/.320 & .105/.246/.333 & .509/.734/.788 & .339/.600/.723 \\
\textbf{BioBERT (NCE)} & .198/.374/.455 & \textbf{.123/.298}/.456 & .467/.686/.776 & .415/.692/.831 \\ 
\midrule
\textbf{OSLAT} & .224/.563/.713 & .018/.193/\textbf{.491} & \textbf{.546/.865/.943} & \textbf{.554/.877/.954} \\
\textbf{OSLAT (CD)} & .238/.450/.577 & \textbf{.123}/.193/.351 & .510/.778/.858 & .308/.646/.785 \\
\textbf{OSLAT (NP)} & .001/.016/.028 & .000/.000/.000 & .004/.009/.019 & .015/.015/.015  \\
\textbf{OSLAT (No LA)} & .041/.071/.105 & 035/.053/.070 & .070/.189/.271 & .015/.138/.246 \\
\bottomrule
\end{tabular}
\caption{Results for entity linking on both datasets, broken down by spans, and evaluated using top-$k$ accuracy (@1, @5, @10).} 
\label{tab:results-linking}
\vspace{-.5em}
\end{table*}

\section{Task 2 Results: Entity Linking}
\label{sec:entity-linking-result}

In this section, we describe the experiments
for the task of entity linking using
the approach described in \S\ref{sec:classifier-entity-linking}.

\subsection{Set-up}
For entity linking, we cache the label representation for all entities $e \in \mathcal{E}$, before computing the retrieval score between all text-entity pairs. In practice, the retrieval score for each pair $(x_t, e_t) \in \mathcal{D}$ is computed as the max over all of $e_t$'s synonyms $P(t)$, s.t. $s(x_t, e_t) = \argmax_{p \in P(t)}~s(x_t, p)$.

\subsection{Metrics} 
To evaluate our entity linker, we use the top-$k$ accuracy with $k=1, 5, 10$. Specifically, for each ground-truth entity mentioned in the target text, we check if the text-entity pair has a top-$k$ retrieval score. 

\subsection{Baselines} 
We use the bi-encoder retrieval architecture as a baseline, where we train BioBERT using the approach in \citet{liu-etal-2021-self}. We first align the entity synonyms in the embedding space, before using the similarity between the target text and entity representation
as the retrieval score.

We compare the following baselines:
\begin{enumerate}
    \item \textbf{Fuzzy-Match.} Similar to \S~\ref{sec:span-baselines}. We use Levenshtein Distance to score.
    \item \textbf{BioBERT (Unsup)}. BioBERT bi-encoder directly as an unsupervised linker
  \item \textbf{BioBERT (MS)}. BioBERT bi-encoder trained with Multi-Similarity (MS) loss. \citep{wang2019multi, liu-etal-2021-self}
    \item \textbf{BioBERT (NCE)}. BioBERT bi-encoder trained with Noise Contrastive Estimation (InfoNCE) loss \citep{chen2020simple, khosla2020supervised, gao-etal-2021-simcse}.
    \item \textbf{OSLAT (NP).} Ablation without self-alignment pertaining (no stage 1).
    \item \textbf{OSLAT (No LA).} Ablation without label attention training (no stage 2).
    \item \textbf{OSLAT (CD).} Cross-dataset evaluation (i.e. RFE model evaluated on hNLP and vice versa). 
\end{enumerate}

\subsection{Results}
\autoref{tab:results-linking} shows experiment results on entity linking. OSLAT outperforms all baselines on the RFE dataset, while lagging behind \textbf{BioBERT (NCE)} for @1 and @5 on the disjoint subset of hNLP. We hypothesize that the linking classifier might be struggling with the higher number of entities in the hNLP test set 
(See \autoref{tab:data_stats}).
As the number of entities increases, there exists a higher chance for semantically similar entities (\eg~back pain vs chronic back pain). Since the classifier only has access to the entity span representation $c^{(x_t, e)}$, which is a weighted sum of the encoder hidden states, it can be difficult to distinguish similar entities without explicitly mining for negatives. BioBERT bi-encoder retriever, on the other hand, has the full contextualized representation of the target text, and therefore may be able to better disambiguate similar entities if the discriminating context can be captured. We include the results for other encoder baselines in Appendix \ref{sec:entity-linking-full}.

\section{Discussion}
\label{sec:discussion}
We propose OSLAT -- a new architecture for entity span extraction and linking. OSLAT augments the standard label attention transformer architecture to allow an open set of labels.  We also introduce a two-step training procedure including a modified supervised contrastive loss function, which we call the synonym-supervised NT-Xent loss.

In our experiments, we show that the two-step pretraining is critical, as  both span extraction (\S~\ref{sec:span-extract}) and entity linking (\S~\ref{sec:entity-linking-result}) task fail to be learned without the self-alignment pretraining of the encoder. This is because without pretraining, the encoder cannot meaningfully distinguish between the synonym and non-synonym representations of entities. 

Through our detailed experiments, we first show that OSLAT implicitly learns to perform span extraction despite being trained only on text-level labels (without any span annotations). In entity linking, we also find that OSLAT outperforms other bi-encoder retrieval architectures. Most importantly, OSLAT performs well on entities mentioned in disjoint spans, which are very common in the colloquial language generated by patients. 


\paragraph*{Ethics} This work was done as part of a quality improvement activity as defined in 45CFR §46.104(d)(4)(iii) -- secondary research for which consent is not required for the purposes of ``health care operations''. In the ``RFE'' dataset, all ground truth annotations were performed by medical professionals who are full-time employees of the company.

\bibliography{li22}

\clearpage
\counterwithin{figure}{section}
\counterwithin{table}{section}
\appendix
\section{Dataset Statistics}

\subsection{RFE Demographics}
\label{sec:rfe-details}
The distribution of biological sexes in the dataset is 75\% female and 25\% male, the distribution of ages is 74\% below 30 years old, 20\% between 30 and 50 years old, and 6\% above 50 years old. This distribution is not a random sample representative of the overall practice’s population, but rather comes from a mixture of random samples drawn from two distinct times, and also from an active learning experiment for a different project.

\subsection{Comparison}
\label{sec:dataset_comparison}
We found that there is a significant difference between the entity sets in both datasets (roughly ~85\%  from hNLP to RFE and  ~69\% from RFE to hNLP), although hNLP has twice the number of entities as the RFE dataset. We attribute the difference between the two datasets to their source; while RFE is derived from a telemedicine practice, hNLP is built from doctor's notes from in-patient settings. Second, only a tiny fraction of \unseen~  entities in one dataset is \seen~ in the other. This gives the assurance that when we evaluate the cross-domain task, we do not provide undue advantage to the model trained on the other dataset just because these \unseen~ entities are known to the other dataset.

\begin{table}[ht!]
\centering
\small
\begin{tabular}{cccccccc}
\toprule
\multicolumn{2}{c}{\multirow{2}{*}{}}                                         & \multicolumn{3}{c}{\textbf{RFE}}                  & \multicolumn{3}{c}{\textbf{hNLP}} \\
\multicolumn{2}{c}{}                                                          & S & U & D                 & S & U & D \\ \hline
\multirow{2}{*}{\textbf{RFE}}                      & \multicolumn{1}{c|}{S }   & 1    &   0     & \multicolumn{1}{c|}{0}    & .23  & .05    & .72      \\
                                          & \multicolumn{1}{c|}{U}  &    0  & 1      & \multicolumn{1}{c|}{0}    & .09  & .24    & .67      \\ \hline
\multicolumn{1}{c}{\multirow{2}{*}{\textbf{hNLP} }} & \multicolumn{1}{c|}{S}  & .10  & .02    & \multicolumn{1}{c|}{.88} & 1    &   0     &    0      \\
\multicolumn{1}{c}{}                      & \multicolumn{1}{c|}{U} & .12  & .04    & \multicolumn{1}{c|}{.84} &   0   & 1      &      0    \\ \bottomrule
\end{tabular}
\caption{Comparison of entities overlap between the two datasets.}
\label{tab:concepts_overlap}
\vspace{-1em}
\end{table}

\label{sec:comparison-details}
In \autoref{tab:concepts_overlap}, we quantitatively compare the overlap of entities between the datasets and make two observations.  
For each dataset (represented by rows), we present the number of entities in the \seen~training set (S), and in the \unseen~open set (U). In the columns corresponding to the other dataset,  we provide the distribution of the occurrence of these entities in their \seen~(S) and \unseen~(U) concept distribution. The last column (D) corresponds to the proportion of concepts not represented in the other dataset (disjoint).

\begin{table}[h]
    \centering
    \begin{tabular}{l}
    \toprule
    \textbf{RFE} \\
    \midrule
     pregnancy, headache, dysuria, cough, \\ abdominal pain, nausea, throat pain, \\
     UTI, delayed menstruation, \\
     vaginal pruritus, vaginal spotting, fever, \\ crampy abdominal pain, fatigue, vomiting \\
     \midrule
     \textbf{hNLP} \\
     \midrule
     systemic arterial hypertension, edema, \\
     chest pain, coronary artery disease, pain, \\ dyspnea, atrial fibrillation, heart failure, \\
     nausea, vomiting, bleeding, \\
     intracerebral hemorrhage, pneumonia, \\
     cyanosis, diabetes mellitus \\
      \bottomrule
    \end{tabular}
    \caption{Top 15 most frequent entities found in the two datasets. }
    \label{tab:freq_concepts}
    \vspace{-1em}
\end{table}

This is also evident when we look at the top frequent entities from these two datasets in \autoref{tab:freq_concepts} where hNLP focuses on more severe health issues (such as heart-related)  that require hospitalization while RFE dataset focuses on non-urgent primary care services. However, they also share entities such as ``vomiting.''

We also found that only a tiny fraction of \unseen~  entities in one dataset is \seen~ in the other. This gives the assurance that when we evaluate the cross-domain task (\S~\ref{sec:crossdomain-results}) we do not provide undue advantage to the model trained on the other dataset just because these \unseen~ entities are known to the other dataset. Note that we did not intentionally construct the datasets this way and this result is a natural consequence of the significant difference in the vocabulary of the two datasets.

\section{Training Hyperparameters}
\label{sec:training-hyperparameters}
For both self-alignment pretraining (\S~\ref{sec:approach_pretrain}) and label attention training (\S~\ref{sec:lbltransformer}), we use the ADAM optimizer \citep{kingma2014adam} with exponential decay after 1/10 of total steps and an effective batch size of $32$. For self-alignment pretraining, we train the model for a total of $20$ epochs with a learning rate of $2e-3$ and the number of negatives set to $50$. For label attention training, we train for a total $10$ epochs with a learning rate of $2e-4$ with the number of negatives set to $100$. We set temperature $\tau$ to $0.07$ based on the settings reported by \citet{khosla2020supervised}. For training the classifier, we also train for a total of $10$ epochs with a learning rate of $2e-4$ and the number of negatives set to $100$. We follow the hyperparameters settings described in \citet{lin2017focal}, where $\alpha=0.25$ and $\gamma = 2.0$

\section{Effects of Self-Alignment Pretraining}
\label{sec:entity-density}

\begin{figure}[ht!]
\centering
\subfigure[Before self-alignment pretraining]{%
\includegraphics[width=0.4\textwidth]{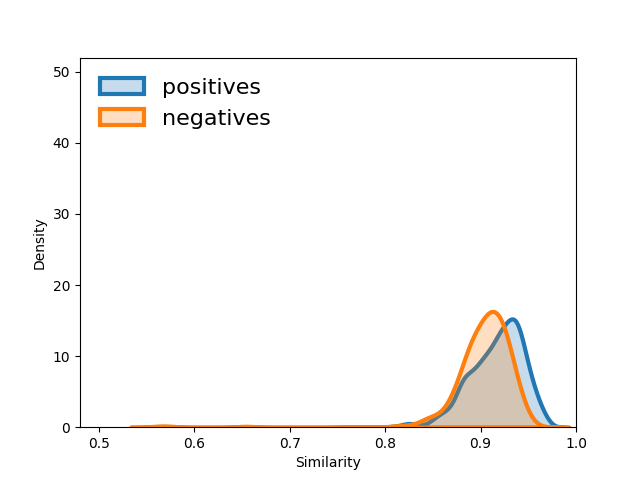}}%
\qquad
\subfigure[After self-alignment pretraining]{%
\includegraphics[width=0.4\textwidth]{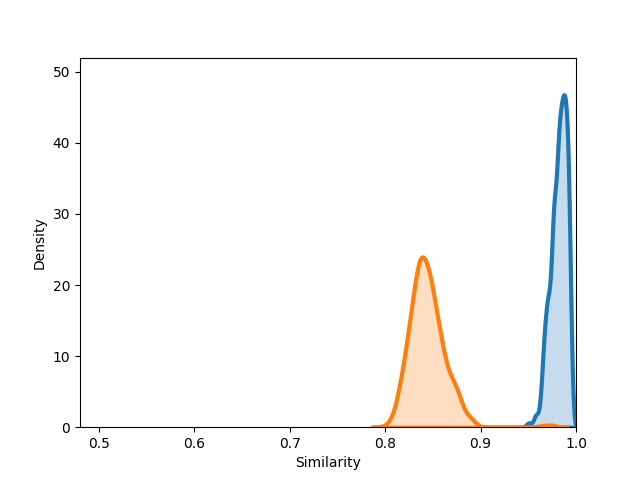}}%
\caption{Density plot of similarities between $1000$ positive and negative entity pairs randomly sampled from  $\mathcal{E}_\seen$ (RFE). 
}
\label{fig:entity_density}
\vspace{-2em}
\end{figure}

To visualize the decrease in representational anisotropy, we plot the similarity between $1000$ positive (synonyms) and negative (non-synonyms) entity pairs randomly sampled from  $\mathcal{E}_\seen$ (RFE). From \autoref{fig:entity_density}, we see that before pretraining (a) the encoder could not differentiate representations of entity synonyms from non-synonyms, while after the pretraining (b), there is a dramatic shift that fully separates synonyms from non-synonyms. 

\section{Detailed metrics breakdown}
\label{apdx:metrics}

\begin{table*}[ht!]
\small
\begin{tabular}{|c|c|c|c|c|c|c|}
\hline
\textbf{Dataset}                                                                         & \textbf{Entities}       & \textbf{\begin{tabular}[c]{@{}c@{}}Metric\\ (Micro)\end{tabular}} & \textbf{\begin{tabular}[c]{@{}c@{}}OSLAT\\ (RFE)\end{tabular}} & \textbf{\begin{tabular}[c]{@{}c@{}}OSLAT\\ (hNLP)\end{tabular}} & \textbf{Rule-Based} & \textbf{Fuzzy}     \\ \hline
\multirow{6}{*}{\textbf{\begin{tabular}[c]{@{}c@{}}RFE\\ Continuous-Span\end{tabular}}}  & \multirow{2}{*}{seen}   & Precision                                                         & 0.69±0.01                                                      & 0.62±0.01                                                       & \multirow{4}{*}{-}  & \multirow{4}{*}{-} \\ \cline{3-5}
                                                                                         &                         & Recall                                                            & 0.65±0.00                                                      & 0.69±0.01                                                       &                     &                    \\ \cline{2-5}
                                                                                         & \multirow{2}{*}{unseen} & Precision                                                         & 0.59±0.01                                                      & 0.53±0.01                                                       &                     &                    \\ \cline{3-5}
                                                                                         &                         & Recall                                                            & 0.59±0.01                                                      & 0.50±0.01                                                       &                     &                    \\ \cline{2-7} 
                                                                                         & \multirow{2}{*}{all}    & Precision                                                         & 0.67±0.01                                                      & 0.57±0.01                                                       & \textbf{0.98}       & 0.90               \\ \cline{3-7} 
                                                                                         &                         & Recall                                                            & \textbf{0.64±0.00}                                             & 0.58±0.01                                                       & 0.38                & 0.21               \\ \hline
\multirow{6}{*}{\textbf{\begin{tabular}[c]{@{}c@{}}RFE\\ Disjoint-Span\end{tabular}}}    & \multirow{2}{*}{seen}   & Precision                                                         & 0.61±0.01                                                      & 0.60±0.01                                                       & \multirow{4}{*}{-}  & \multirow{4}{*}{-} \\ \cline{3-5}
                                                                                         &                         & Recall                                                            & 0.51±0.02                                                      & 0.54±0.01                                                       &                     &                    \\ \cline{2-5}
                                                                                         & \multirow{2}{*}{unseen} & Precision                                                         & 0.62±0.02                                                      & 0.51±0.01                                                       &                     &                    \\ \cline{3-5}
                                                                                         &                         & Recall                                                            & 0.58±0.01                                                      & 0.38±0.01                                                       &                     &                    \\ \cline{2-7} 
                                                                                         & \multirow{2}{*}{all}    & Precision                                                         & 0.61±0.01                                                      & 0.54±0.01                                                       & \textbf{0.95}       & 0.64               \\ \cline{3-7} 
                                                                                         &                         & Recall                                                            & \textbf{0.53±0.02}                                             & 0.44±0.01                                                       & 0.12                & 0.12               \\ \hline
\multirow{6}{*}{\textbf{\begin{tabular}[c]{@{}c@{}}hNLP\\ Continuous-Span\end{tabular}}} & \multirow{2}{*}{seen}   & Precision                                                         & 0.67±0.00                                                      & 0.66±0.02                                                       & \multirow{4}{*}{-}  & \multirow{4}{*}{-} \\ \cline{3-5}
                                                                                         &                         & Recall                                                            & 0.97±0.00                                                      & 0.92±0.01                                                       &                     &                    \\ \cline{2-5}
                                                                                         & \multirow{2}{*}{unseen} & Precision                                                         & 0.52±0.01                                                      & 0.61±0.01                                                       &                     &                    \\ \cline{3-5}
                                                                                         &                         & Recall                                                            & 0.88±0.01                                                      & 0.90±0.00                                                       &                     &                    \\ \cline{2-7} 
                                                                                         & \multirow{2}{*}{all}    & Precision                                                         & 0.57±0.01                                                      & 0.61±0.01                                                       & \textbf{0.98}       & 0.70               \\ \cline{3-7} 
                                                                                         &                         & Recall                                                            & \textbf{0.91±0.01}                                             & 0.90±0.01                                                       & 0.64                & 0.89               \\ \hline
\multirow{6}{*}{\textbf{\begin{tabular}[c]{@{}c@{}}hNLP\\ Disjoint-Span\end{tabular}}}   & \multirow{2}{*}{seen}   & Precision                                                         & 0.47±0.02                                                      & \multirow{2}{*}{-}                                              & \multirow{4}{*}{-}  & \multirow{4}{*}{-} \\ \cline{3-4}
                                                                                         &                         & Recall                                                            & 0.71±0.02                                                      &                                                                 &                     &                    \\ \cline{2-5}
                                                                                         & \multirow{2}{*}{unseen} & Precision                                                         & 0.43±0.02                                                      & 0.45±0.01                                                       &                     &                    \\ \cline{3-5}
                                                                                         &                         & Recall                                                            & 0.45±0.02                                                      & 0.47±0.01                                                       &                     &                    \\ \cline{2-7} 
                                                                                         & \multirow{2}{*}{all}    & Precision                                                         & 0.44±0.02                                                      & 0.45±0.01                                                       & \textbf{0.72}       & 0.49               \\ \cline{3-7} 
                                                                                         &                         & Recall                                                            & \textbf{0.51±0.02}                                             & 0.47±0.01                                                       & 0.33                & 0.32               \\ \hline
\end{tabular}
\vspace{-.5em}
\caption{The breakdown of the micro-precision and recall performance on both datasets. We report the results for both of our models and the two baseline methods along with the standard deviation across 5 random seeds.}
\label{tab:precision_recall}
\vspace{-1.5em}
\end{table*}

In this section, we provide a detailed breakdown of the results from \autoref{tab:results-main}, where we discuss the recall-precision trade-off between our models and the two baseline methods. From the results in \autoref{tab:precision_recall}, we see that while the RFE trained OSLAT achieved higher recall against both baseline methods, the rule-based model achieved higher precision across all datasets, with near-perfect precision for contiguous span entities. This is expected since the rule-based model has access to the ground-truth entity, the predictions it makes almost always exactly match with the entity or one of its synonyms. On the other hand, OSLAT can extract implicitly mentioned entities and disjoint-spans based on semantic similarity, resulting in a higher recall across all datasets. We leave the exploration of ensembling the two methods as a potential direction for future work. Lastly, it is worth mentioning that the precision and recall trade-off for OSLAT could be manually adjusted by tuning the prediction threshold of the attention scores. However, due to the limited size of our training set, we only report the performance for a fixed threshold (0.05).

\section{Extended Results for Entity Linking}
\label{sec:entity-linking-full}

\vspace{-.5em}

In \autoref{tab:results-linking-full}, we report the extended results for entity linking including the baselines of SAP-BERT \citep{liu-etal-2021-self} and PubMedBERT \citep{pubmedbert-2021}. However, they are not directly comparable with \textbf{OSLAT}, since our model is based on BioBERT \citep{lee2020biobert}. We leave the experiments of OSLAT with other encoders as future work. Lastly, we also include the results using the ground-truth entity spans during inference (\textbf{OSLAT (GT)}). This is done by mean-pooling over the hidden states associated with the entity mention spans (rather than using the attention scores), before applying the binary classifier for prediction.

\begin{table*}[ht!]
\small
\begin{tabular}{@{}lcccc@{}}
\toprule
\multicolumn{1}{c}{\multirow{2}{*}{Dataset}} & \multicolumn{2}{c}{\textbf{hNLP}} & \multicolumn{2}{c}{\textbf{RFE}} \\
\multicolumn{1}{c}{} & Contiguous-Span & Disjoint-Span & Contiguous-Span & Disjoint-Span \\
Accuracy & @1/@5/@10 & @1/@5/@10 & @1/@5/@10 & @1/@5/@10 \\ 
\midrule
\textbf{BioBERT (Unsup)} & .080/.134/.168 & .070/.105/.211 & .186/.031/.352 & .077/.154/.185 \\
\textbf{BioBERT (MS)} & .173/.280/.320 & .105/.246/.333 & .509/.734/.788 & .339/.600/.723 \\
\textbf{BioBERT (NCE)} & .198/.374/.455 & .123/.298/.456 & .467/.686/.776 & .415/.692/.831 \\ 
\midrule
\textbf{SAP-BERT (Unsup)} & .184/.297/.358 & .193/.333/.404 & .269/.403/.470 & .108/.292/.354 \\
\textbf{SAP-BERT (MS)} & .157/.234/.271 & .088/.193/.211 & .477/.705/.769 & .354/.585/.615  \\
\textbf{SAP-BERT (NCE)} & .206/.359/.435 & .123/.351/.474 & .197/.304/.351 & .092/.185/.262 \\ 
\midrule
\textbf{PubMedBERT (Unsup)} & .080/.134/.168 & .070/.105/.211 & .144/.206/.234 & .077/.139/.154\\
\textbf{PubMedBERT (MS)} & .197/.313/.363 & .105/.246/.316 & .523/.751/.820 & .354/.692/.800\\
\textbf{PubMedBERT (NCE)} & .201/.379/.494 & .175/.351/.561 & .318/.447/.517 & .200/.431/.539\\ 
\midrule
\textbf{OSLAT} & .224/.563/.713 & .018/.193/.491 & .546/\textbf{.865}/\textbf{.943} & \textbf{.554}/\textbf{.877}/\textbf{.954} \\
\textbf{OSLAT (CD)} & .238/.450/.577 & .123/.193/.351 & .510/.778/.858 & .308/.646/.785 \\
\textbf{OSLAT (NP)} & .001/.016/.028 & .000/.000/.000 & .004/.009/.019 & .015/.015/.015  \\
\textbf{OSLAT (No LA)} & .041/.071/.105 & 035/.053/.070 & .070/.189/.271 & .015/.138/.246 \\
\textbf{OSLAT (GT)} & \textbf{.483}/\textbf{.629}/\textbf{.752} & \textbf{.439}/\textbf{.597}/\textbf{.737} & \textbf{.555}/.793/.871 & .462/.785/.908 \\
\bottomrule
\end{tabular}
\caption{Results for entity linking on both datasets, broken down by spans, and evaluated using top-$k$ accuracy (@1, @5, @10).} 
\label{tab:results-linking-full}
\vspace{-2em}
\end{table*}

\end{document}